\documentclass[acmsmall,nonacm]{acmart}
\AtBeginDocument{%
  }

\usepackage{multirow}
\usepackage[table]{xcolor}
\usepackage{booktabs}
\usepackage{wrapfig}

\definecolor{low}{RGB}{198,219,239}      %
\definecolor{medium}{RGB}{158,202,225}   %
\definecolor{high}{RGB}{49,130,189}      %
\definecolor{labelcolor}{RGB}{255,255,255} %

\usepackage{cleveref}

\begin{document}
\raggedbottom

\title{What We are Missing in Multimodal LLM Evaluation?}

\author{Po-han Li}
\authornote{Both authors contributed equally to this research.}
\email{pohanli@utexas.edu}
\orcid{0009-0007-2206-3546}
\author{Shenghui Chen}
\authornotemark[1]
\email{shenghui.chen@utexas.edu}
\author{Sandeep Chinchali}
\email{sandeepc@utexas.edu}
\author{Ufuk Topcu}
\email{utopcu@utexas.edu}
\affiliation{%
  \institution{The University of Texas at Austin}
  \city{Austin}
  \state{Texas}
  \country{USA}
}

\renewcommand{\shortauthors}{Li et al.}

\begin{abstract}
Multimodal large language models (MLLMs) can process diverse inputs, \textit{e.g.}, text, images, audio, and video, and generate textual responses. While their capabilities have advanced rapidly, evaluation of such models has not kept pace. Most existing evaluation benchmarks are limited to isolated tasks and reveal little about whether a model integrates information across modalities.
We examine current means for evaluating MLLMs and review the existing benchmark taxonomy to identify gaps, including temporal-spatial coherence, physical world understanding, multimodal consistency, and selective attention. Addressing these gaps is essential for measuring real progress in multimodal intelligence and exposing capability boundaries.
\end{abstract}



\maketitle

\section{Introduction}
Multimodal large language models (MLLMs), such as Gemini-3.1-Pro and GPT-5, process multiple modalities of input, including text, images, audio, and video, and produce textual responses. These models aim to unify perception and reasoning within a single framework. By combining several modalities, MLLMs perform tasks such as video understanding and analysis of multimedia documents. Despite rapid growth in capabilities, evaluation methods have not kept pace.

\begin{figure}
    \centering
    \includegraphics[width=\linewidth]{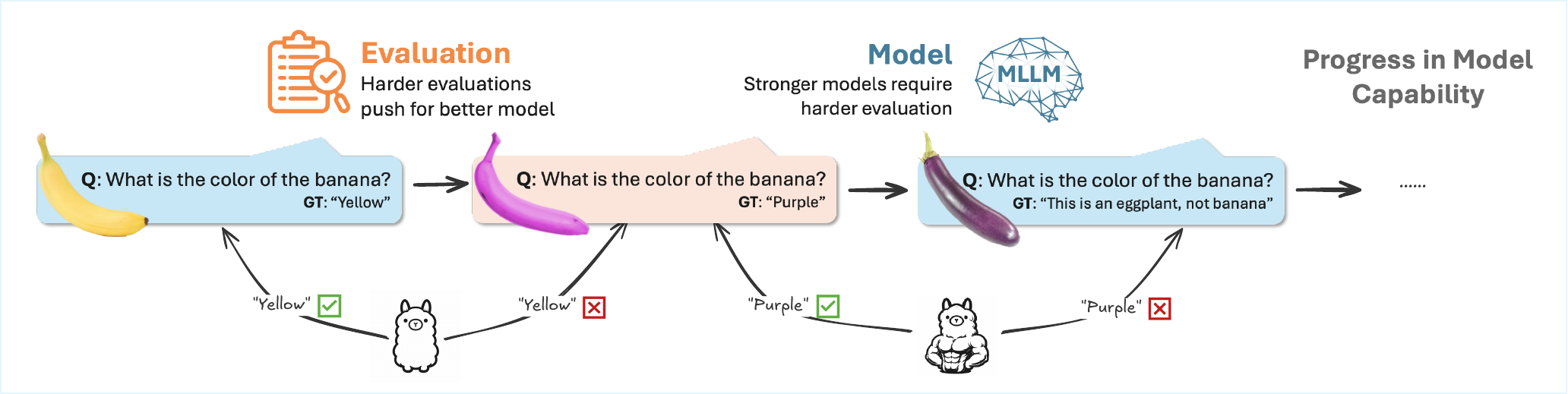}
    \caption{The iterative feedback loop between MLLM development and evaluation. Stronger models demand more challenging evaluations, which in turn reveal latent failure modes.}
    \label{fig:helix}
\end{figure}

Evaluation serves as a critical component in the iterative development cycle of MLLMs, as illustrated in \Cref{fig:helix}. Rather than acting solely as a diagnostic tool, evaluation forms a feedback loop with model development: stronger models demand more challenging benchmarks, while harder evaluations in turn drive the development of more capable models. Evaluation helps identify capability boundaries and failure modes, such as cross-modal discrepancy, where MLLMs process each modality correctly but fail to combine them into a coherent output. A well-known example in \Cref{fig:helix} illustrates this limitation: when asked ``what is the color of the banana?'' given an image of a banana, earlier MLLMs frequently answered yellow, reflecting the dominant color in their pretraining data \cite{kv2020reducing}. This over-reliance on textual priors would go unnoticed if evaluations never tested bananas of other colors.

Unlike unimodal large language models (LLMs), MLLMs must connect multiple sensory inputs to textual reasoning, requiring a framework that verifies cross-modal alignment.
Most existing benchmarks focus on isolated tasks, such as recognizing objects in images or answering short questions. While success on these benchmarks indicates proficiency on specific subtasks, it reveals little about whether a model truly integrates information across modalities. 

Real-world deployment exposes multimodal failures that standard evaluation often misses. Models may misattribute actions in videos or overlook key details in complex documents. Such errors introduce safety and accountability risks, especially in autonomous driving, medical imaging, and financial analysis. 

Therefore, new evaluations become necessary.
Multimodal evaluation must align perceptual signals with logical reasoning to uncover failure modes that text-only benchmarks miss. It should assess whether models integrate multiple modalities, preserve temporal-spatial coherence, and generate reliable outputs under realistic conditions. From a human-centered perspective, evaluation should measure not just benchmark accuracy, but also reliability under time pressure and ambiguous cross-modal evidence. Key capabilities often insufficiently assessed include temporal-spatial coherence, physical world understanding, multimodal consistency, and selective attention.

\section{Why Current Evaluations Fail in the Real World}

As MLLMs move from experimental tools to real-world applications such as medical diagnostics, autonomous systems, and financial analysis, the cost of error rises sharply. A hallucination reflects a failure to ground information across modalities—for example, misreading temporal events in video or missing a link between an audio alert and a visual signal. Real-world deployment leaves little margin for error and therefore requires evaluation to be more rigorous.

Part of the current reliability crisis stems from the limitations of traditional \textit{static benchmarks}. Once used to track progress, these datasets now often fail to reveal how MLLMs integrate information across modalities.
Two key issues contribute to this problem are data leakage and benchmark contamination.
Data leakage occurs when benchmark examples overlap with model training data, often because training and evaluation sets are both scraped from the same ubiquitous internet sources. Hence, MLLMs may appear to reason through a complex task when it is actually recalling a specific data pair from their pretraining.
Benchmark contamination arises when ground-truth labels are incorrect or the provided context is insufficient or contradictory. It worsens with dense, multifaceted data, such as video-audio synchronization or complex technical diagrams. It compromises the validity of evaluation by inflating reported scores and obscuring true generalization, making it difficult to separate memorization from genuine multimodal reasoning.

To address the limitations of static benchmarks, the community has begun to explore interactive, in-the-wild evaluation platforms. Systems such as LMArena compare models through large-scale pairwise voting by human users, allowing evaluation through open-ended conversations rather than fixed datasets~\cite{chiang2024chatbot}. Recent extensions have begun incorporating MLLMs, though remain largely limited to static images. While these platforms provide valuable insight into real-world usage patterns, they also introduce new biases. Human raters often prefer responses that are fluent, natural, or lengthy rather than strictly accurate or informative. In addition, model developers may fine-tune systems specifically to improve rankings and again couple model development to leaderboard performance rather than underlying capability.

The result is a paradoxical landscape where the industry sees a constant stream of new state-of-the-art rankings that consistently fail to translate into practical utility for users. It leads to a capability gap characterized by a discrepancy between impressive leaderboard scores and the often inconsistent user experience. To bridge this gap, evaluation must evolve beyond simple pattern learning and move toward measuring functional modality integration. 
We must ensure that a model's high performance reflects an ability to synthesize various sensory cues, such as visual signals, auditory input, or textual instructions, while applying logical constraints. 
Evaluation should also capture tangible benefits for users, including time saved, money saved, or effort reduced.

\begin{figure}
    \centering
    \includegraphics[width=\linewidth]{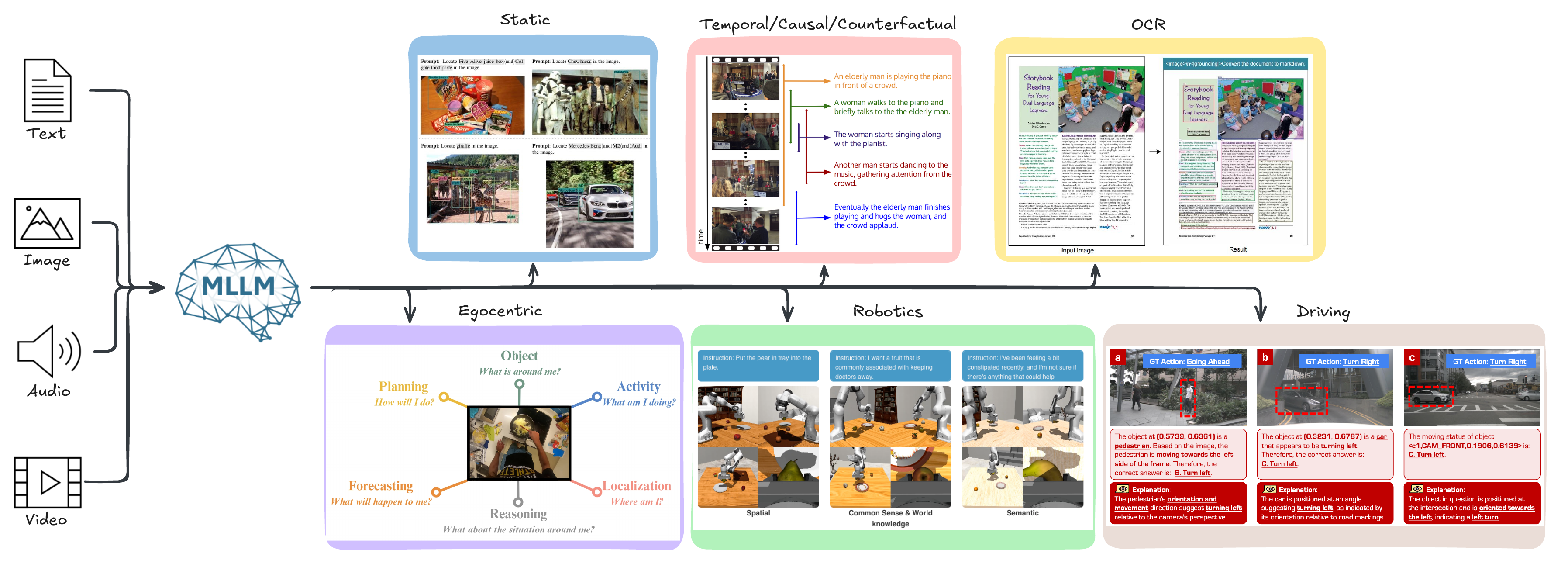}
    \caption{
    MLLMs leverage multimodal integration (text, image, audio, and video) to produce textual outputs for various tasks.
    Example credit by \cite{xiao2024florence,krishna2017dense,wei2025deepseek,Cheng_2024_CVPR,vlabench,xie2025drivebench}.}
    \label{fig:taxonomy}
\end{figure}

\section{Multimodal Benchmarking Taxonomies}
Current evaluation remains dominated by static, image-centric benchmarks that test perception and visual question answering (as shown in \Cref{fig:taxonomy}). While these tasks can reach high levels of difficulty, they largely ignore temporal dynamics and cross-modal interactions.

Video benchmarks extend evaluation to extended sequences, measuring object tracking and causal reasoning over time. However, they often emphasize limited visual moments and omit audio, reducing multimodal understanding to a primarily visual task. Similarly, OCR benchmarks focus on structured documents, requiring models to parse layouts and extract semantic meaning, but remain confined to static inputs.

More realistic evaluation emerges in embodied and egocentric settings. Robotics benchmarks, such as VLABench \cite{vlabench}, assess whether models can translate perception into action, requiring spatial grounding and long-horizon reasoning. Egocentric benchmarks further test first-person understanding of human intent and interactions under occlusion and motion. Despite their promise, these settings remain limited in scope and standardization.

Across these benchmarks, a common limitation persists: they evaluate capabilities in isolation rather than testing whether models integrate information into a coherent, grounded understanding. This is compounded by an over-reliance on multiple-choice formats and pre-defined gold standards. As noted in OpenAI's report \cite{kalai2025language}, MLLMs (and LLMs) exhibit a ``forced-choice" bias, tending to guess answers rather than acknowledge visual ambiguity or internal uncertainty. It exposes an evaluation gap where ground truth is often absent or fluid, particularly in domains such as reinforcement learning, autonomous driving, and human-aligned tasks. 
Expanding evaluation formats, for example, by correlating model performance with physical-world tasks \cite{vlabench} or human task effectiveness \cite{vibe}, allows for more sophisticated measures of performance.

\section{Multimodal Testbed Capabilities for New Evaluations}

\begin{table}[t]
    \centering
    \resizebox{0.75\linewidth}{!}{
        \begin{tabular}{lcccccc}
        \toprule
        \multirow{2}{*}{Capability} & \multicolumn{6}{c}{Benchmark} \\
        \cmidrule(lr){2-7}
        & Static & OCR & Temporal/Causal/Counterfactual & Egocentric & Robotics & Driving \\
        \midrule
        Temporal Coherence & 
        \cellcolor{low}L & 
        \cellcolor{low}L & 
        \cellcolor{high}H & 
        \cellcolor{high}H & 
        \cellcolor{medium}M &
        \cellcolor{high}H \\
        
        Spatial Reasoning & 
        \cellcolor{medium}M & 
        \cellcolor{medium}M & 
        \cellcolor{medium}M & 
        \cellcolor{high}H & 
        \cellcolor{high}H &
        \cellcolor{high}H \\
        
        Physical World Modeling & 
        \cellcolor{low}L & 
        \cellcolor{low}L & 
        \cellcolor{high}H &
        \cellcolor{medium}M & 
        \cellcolor{high}H &
        \cellcolor{high}H \\
        
        Multimodal Integration & 
        \cellcolor{low}L & 
        \cellcolor{medium}M & 
        \cellcolor{medium}M & 
        \cellcolor{high}H & 
        \cellcolor{high}H &
        \cellcolor{high}H \\
        
        Saliency/Attention & 
        \cellcolor{medium}M & 
        \cellcolor{medium}M & 
        \cellcolor{low}L & 
        \cellcolor{high}H & 
        \cellcolor{high}H &
        \cellcolor{medium}M \\
        
        \bottomrule
        \end{tabular}
    }
    \caption{MLLM capability coverage heatmap (L: low, M: medium, H: high).}
    \label{tab:cap}
    \vspace{-20pt}
\end{table}

The transition of evaluation formats necessitates a shift in how we define a model's capability. New evaluations must measure the integration of multimodalities into a coherent internal representation of the world. The capabilities define the boundary between models that merely process pixels and those that possess an understanding of the world similar to human perception. Besides question-answering benchmarks, we require interactive evaluations that go beyond ground-truth answers, testing response and adaptation in dynamic scenarios.
We highlight several capabilities that remain insufficiently assessed and summarize their relationship to existing evaluations in Table 1.

\paragraph{\textbf{Temporal and Spatial Coherence}}
These capabilities are grounded in the dual requirement of temporal and spatial coherence. Temporal analysis evaluates a model's ability to maintain narrative continuity over time, requiring an understanding of sequence and causality that spans minutes of input. It involves recognizing an action and understanding the "before and after" that defines an event. Spatial analysis tests a model’s understanding of 3D geometry and object positioning, requiring geometric consistency despite changes in camera or environment.

\paragraph{\textbf{Physical World Modeling}}
True multimodal ability requires modeling of physical laws, often tested through changing viewpoints and object interactions. Benchmarks should evaluate "different person views" to test object permanence, requiring models to track items across disparate viewpoints or during periods of full occlusion. This capability is paired with interactive object tracking, where models must infer how objects behave in the 3D space when acted on by external forces. Beyond tracking, evaluations should also assess a model's physical understanding, specifically depth, velocity, and collision trajectories. A critical metric is causal forecasting: the ability to predict the rational outcome of a physical sequence, \textit{e.g.}, a glass falling onto a hard surface will likely shatter.

\paragraph{\textbf{Multimodal Integration}}
The most advanced stage of evaluation focuses on the synergy between modalities, requiring models to maintain a unified internal state across various sensory inputs. Multimodal processing enforces strict alignment and consistency, penalizing models that generate contradictory outputs. For example, a model describing the sound of a glass breaking when the visual evidence shows it remained intact. This cross-modal verification is essential for ensuring that the MLLMs are not merely hallucinating disparate details but are instead synthesizing a singular, grounded understanding from its inputs.

\paragraph{\textbf{Saliency and Selective Attention}}
Saliency and selective attention measure whether a model prioritizes task-relevant signals and suppresses irrelevant ones across modalities. Reliable systems highlight critical cues, such as abrupt motion, speaker shifts, or anomalous readings, rather than weighting all inputs equally. Perturbation, distractor, and attribution tests evaluate whether predictions respond correctly when key signals change. It is vital, since many real-world failures stem from focusing on the wrong signal, not missing perception. Benchmarks that ignore attention may reward fragile models that succeed for the wrong reasons.

\section{Conclusion}

The growing gap between leaderboard scores and real-world performance underscores the need to rethink how we assess model capability. 
Current benchmarks drove rapid progress by providing clear targets, but are increasingly subject to \textit{Goodhart's Law}: when a measure becomes a target, it ceases to be a good measure. For example, in economics, over-optimizing key performance indicators can distort decision-making. Metrics that were once useful proxies degrade under sustained optimization pressure.

Potential solutions involve designing multi-metric evaluation frameworks that go beyond a single scalar value to capture trade-offs across multiple dimensions. They also include keeping portions of the evaluation confidential to prevent overfitting and leaderboard gaming. Additionally, periodically refreshing datasets' content helps reflect novel multimodal scenarios.

\end{document}